\documentclass[runningheads]{llncs}
\usepackage[T1]{fontenc}
\usepackage{arydshln}
\usepackage{graphicx}
\usepackage{hyperref}
\usepackage{color}
\usepackage{amssymb}

\begin{document}
\title{ReCAP: Recursive Cross Attention Network for Pseudo-Label Generation in Robotic Surgical Skill Assessment}
%
\titlerunning{ReCAP: Pseudo-Label Generation for Skill Assessment}
\author{Julien Quarez$^{1}$, Marc Modat$^{1}$,
 Sebastien Ourselin$^{1}$, Jonathan Shapey$^{1,2}$, Alejandro Granados$^{1}$
\thanks{Acknowledgments. This work was supported by core funding from the Wellcome/EPSRC Centre for Medical Engineering [WT203148/Z/16/Z]. For the purpose of Open Access, the Author has applied a CC BY public copyright license to any Author Accepted Manuscript version arising from this submission}
\institute{$^{1}$School of Biomedical Engineering and Imaging Sciences, King's College London $^{2}$Neurosurgery, King's College Hospital%
}}
\authorrunning{J. Quarez et al.}

\maketitle              
\begin{abstract}
In surgical skill assessment, the Objective Structured Assessments of Technical Skills (OSATS) and Global Rating Scale (GRS) are well-established tools for evaluating surgeons during training. These metrics, along with performance feedback, help surgeons improve and reach practice standards. Recent research on the open-source JIGSAWS dataset, which includes both GRS and OSATS labels, has focused on regressing GRS scores from kinematic data, video, or their combination. However, we argue that regressing GRS alone is limiting, as it aggregates OSATS scores and overlooks clinically meaningful variations during a surgical trial.
To address this, we developed a weakly-supervised recurrent transformer model that tracks a surgeon’s performance throughout a session by mapping hidden states to six OSATS, derived from kinematic data. These OSATS scores are averaged to predict GRS, allowing us to compare our model’s performance against state-of-the-art (SOTA) methods. We report Spearman’s Correlation Coefficients (SCC) demonstrating that our model outperforms SOTA using kinematic data (SCC 0.83-0.88), and matches performance with video-based models.

Our model also surpasses SOTA in most tasks for average OSATS predictions (SCC 0.46-0.70) and specific OSATS (SCC 0.56-0.95). The generation of pseudo-labels at the segment level translates quantitative predictions into qualitative feedback, vital for automated surgical skill assessment pipelines. A senior surgeon validated our model’s outputs, agreeing with 77\% of the weakly-supervised predictions \(p=0.006\).
\end{abstract}

\section{Introduction}
Robotic surgery has rapidly expanded across specialties and autonomy levels. Although evidence on the benefits of robotic-assisted surgery (RAS) is mixed~\cite{Maynou2022}, its use is increasing—for example, in the UK, RAS for radical prostatectomy rose from 5\% in 2006 to 88\% in 2018~\cite{Maynou2022}, though adoption varies widely (e.g., 1.8\% in ENT). A significant barrier to wider adoption is the variability in training across systems and institutions~\cite{rastraining,Ravi2021}. Skill assessment is central to surgical training, enabling evaluation of trainee progress. However, reliance on senior surgeons for feedback limits junior doctors' opportunities~\cite{Khairy2004}. Automated, system-agnostic assessment could overcome this.

The Objective Structured Assessment of Technical Skills (OSATS)~\cite{Martin1997} provides a standardized, Likert-scale framework widely used in RAS. It yields a Global Rating Score (GRS) summarizing multiple skill components. Despite its usefulness, OSATS depends on expert assessors~\cite{Singh2014,HACKNEY202360} and remains time-consuming and subjective. Machine learning (ML) and deep learning (DL) offer promising paths to automate and scale assessment.

Kinematic data is particularly suited for this task, offering standardization, lower computational cost, and system-agnostic features compared to video data \cite{Liu2020,Zago2019,yanik2021deep,Kulik2023}. Recent works focus on regressing GRS or expertise levels \cite{Evangelos,Li,Wang2020,Zia2018,IsmailFawaz2019}, but these high-level scores provide limited clinical insight and still require expert feedback. Efforts to model score changes during procedures \cite{IsmailFawaz2019,Khalid2020,Clearness,Wang2020,Zia2018,Wang2021,Evangelos} face challenges such as increased labeling burden~\cite{Clearness,Wang2020} or insufficient validation~\cite{Zia2018,Wang2021,Evangelos}.

For instance, Wang \textit{et al.}~\cite{Wang2020} used supervised recurrent networks on intermediate GRS scores but depended on granular labels and lacked interpretability~\cite{MARKUS,Kelly2019}. Anastasiou \textit{et al.}~\cite{Evangelos} employed contrastive learning for GRS regression, yet their method lacks actionable feedback. Zia \textit{et al.}~\cite{Zia2018} explored hand-crafted features to link input segments to OSATS, improving interpretability but without directly translating predictions into clinical feedback.

Our work addresses this gap by predicting intermediate OSATS scores during a surgical trial in a weakly-supervised manner, without additional labels. Using a recurrent cross-attention model on kinematic data, we provide segment-level OSATS predictions that offer detailed, actionable insights, advancing automated surgical skill assessment.

We summarize our contributions as follows:
\begin{enumerate}
  \item An objective function enabling recurrent cross-attention models to predict trial-level GRS and OSATS scores alongside granular, segment-level OSATS scores in a weakly-supervised way, outperforming existing kinematics-based methods.
  \item Revisiting kinematic data for task-agnostic modeling that links segment-level OSATS predictions to qualitative feedback.
\end{enumerate}

\section{Methods}
We propose a recurrent model called ReCAP, Recursive Cross-Attention for Pseudo-label generation, where segments of kinematic data are processed into intermediate OSATS scores (Fig. \ref{fig:model}). Those scores are then averaged into trial-level OSATS predictions. Our multi-task model is trained in an end-to-end fashion to output all six OSATS. 
We assess the model's performance on the GRS label by aggregating the individual OSATS predicted scores. 

\subsubsection{Problem Formulation}

An input signal \(X_i \in \mathbb{R}^{D \times T_i}\), of feature size D and length \(T_i\), is divided into equal segments \(x^s_i\) of size \(L\) (\(x^s_i \in \mathbb{R}^{D \times L} \)):  \( \{x^{1}_i, x^{2}_i, \dots, x^{s}_i\} \in X_i \) where \(S_i \)  is the total number of segments, i.e. \(S_i = \frac{T_i}{L}\) for a given signal \(i\). For simplicity, in the rest of this paper, we omit \(i\) and use \(s\) as a subscript to refer to different segments within a trial \(i\), i.e. \(x^s_i \to x_s\).
We fit a function \(F\) to map \(X\) to the label space \(\mathcal{Y}:(F:X \to \vec{Y})\):
\begin{equation}
\vec{Y} = F(x_{1}, x_{2}, \ldots, x_{s}, \ldots x_{S}) \label{eq:1}
\end{equation}
where \(\vec{Y} \in \mathcal{Y} \) is a vector composed of all OSATS.
The GRS is the aggregate of OSATS scores: \(Y =\sum y_n\) where \(y_n\) is the \(n^{th} \) OSATS.

Considering clinical practice where a given score is representative of their average performance through the trial i.e.: \(y_n = \frac{1}{S} \sum y^{n}_{s} \), we rewrite Eq. \ref{eq:1} into Eq. \ref{eq:2}, where \(f_{n}\) maps a segment \(s\) to the \(n^{th}\) OSATS intermediate label (\(x_{s} \to y^{n}_{s}\)), similarly to many-to-many training in recursive training. Note that there is no ground truth for \(y_{s}^{n}\) and we learn \(\hat{y}_{s}^{n}\) in a weakly-supervised manner.
\begin{equation}
    y_n = \frac{1}{S}\sum_{s=1}^{S}f_{n}(x_{s}) \label{eq:2}
\end{equation}

\subsubsection{Model Overview}
\begin{figure*}
    \centering
    \includegraphics[width=1\textwidth]{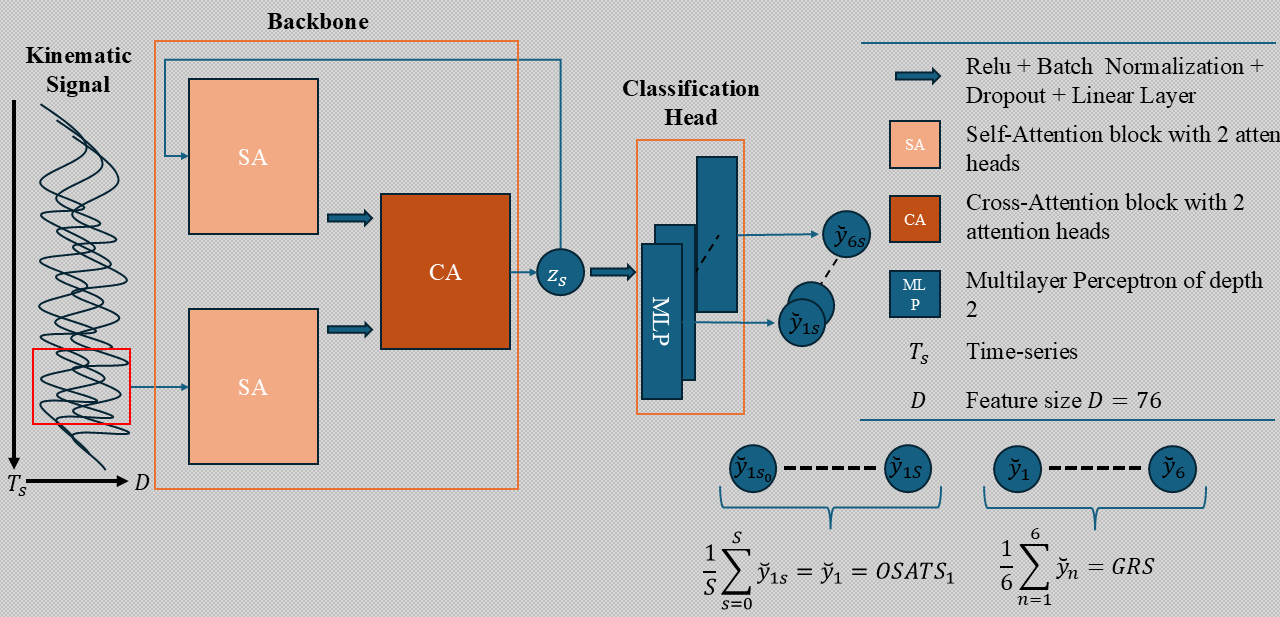}
    \caption{\textbf{ReCAP Architecture Overview}: A kinematic signal is split into segments \(x_s\) of size \(L\) and used as inputs to our backbone \(h\) recurrently. The previous state \(z_{s-1}\) is also passed as an input and fused to \(x_s\) to produce \(z_s\).  This fusion is performed by a fusion module consisting of self- and cross-attention blocks. The current hidden state \(z_s\) is given as an input to six classification heads \(c_n\) to predict the respective OSATS score. \(\hat{y}_{1s}\) to \(\hat{y}_{6s}\) corresponding to the respective OSATS category prediction at segment position \(s\)}
    \label{fig:model}
\end{figure*}

:
Our model recurrently processes segments of a kinematic signal by taking two inputs: the previous hidden state of the recurrent network, \(z_{s-1} \in \mathbb{R}^{D \times L}\), and the current segment-level kinematic signal, \(x_{s}\) (Fig. \ref{fig:model}). 
The two inputs are fused into the current hidden state, \(z_s = h(x_s,z_{s-1})\), through the model backbone \(h\) (Fig. \ref{fig:model}). We initialise \(z_{0}\) as a zero-filled tensor.
Each hidden state is then passed to six classification heads, \(c_n\), giving \(f_n(x_s)= c_n(h(x_s,z_{s-1}))\).
The output of our model is a final \(n^{th}\) OSATS score, the average of all segment-level OSATS predictions:
\begin{equation}
    \hat{y_n} = \frac{1}{S}\sum\limits_{s=0}^{S}c_{n}[h(x_s, z_{s-1})]\ \label{eq:3}
\end{equation}

\noindent \textbf{The backbone} \(h\) is composed of one fusion module (Fig. \ref{fig:model}), where previous temporal information is fused with the current input through a series of multi-head self- and cross-attention blocks.

\noindent \textbf{The classification heads} \(c_n\) are five multilayer perceptron (MLPs) classifying the hidden state \(z_s\) into segment-level OSATS predictions $\hat{y}_{ns} = c_n(z_{s})$.
Each MLP layer consists of batch normalisation, ReLU activation function, and fully connected layers.

\noindent \textbf{Loss}: ReCAP is trained end-to-end using a cross-entropy loss. The loss is applied to the average of the classification head segment predictions $\hat{y_n} = \frac{1}{S}\sum \hat{y}_{ns}$ for a given OSATS category label $y_n$.
 An \(L2\) penalty term weighted by \(\lambda \) is added to regularize the network to help with generalisation. Our final loss is expressed as: 
\begin{equation}
 \mathcal{L} = \sum_{n=0}^{N} CE(\hat{y_n}, y_n) + \lambda * L2  \label{eq:4}
\end{equation}

\subsubsection{Experimental Design}:
 We evaluated our model on the JIGSAWS  dataset \cite{jigsaw}. This dataset consists of video and kinematics data generated by eight clinicians evaluated on three distinct tasks, namely needle passing (NP), suturing (SU), and knot-tying (KT). Altogether, there are 39, 28, and 36 labelled data samples for SU, NP, and KT, respectively. The labels are comprised of six OSATS score (1-5) and one GRS (6-35) the aggregate of all OSATS. It is worth noticing that the dataset is biased towards lower OSATS as the score 5 only appears in the suturing task. The OSATS include 1) respect for tissue, 2) suture/needle handling, 3) time and motion, 4) flow of operation, 5) overall performance, and 6) quality of the final product.

\noindent \textbf{Cross-validation scheme}: Following the JIGSAWS cross-validation framework, we evaluate our method using Leave-One-Supertrial-Out (LOSO) 
whereby the i-th trial performed by the surgeons are left out as the validation set. The cross-validation scheme Leave-One-User-Out (LOUO) is not considered in this work since there is a lack of literature reporting OSATS for kinematic data. Literature \cite{Benmansour2023,Evangelos} on the JIGSAWS dataset seems to indicate that no testing fold is used to assess performance.  

\noindent \textbf{Evaluation}: 
Similar to relevant work \cite{Zia2018,Gao2020,Liu,pan}, we evaluate our method using Spearman's Correlation Coefficient (SCC) \(\rho\) to compare the predicted ranked GRS score with the ground truth: 
\begin{equation}
\rho = 1 - \frac{6\sum d_i^2}{n(n^2 - 1)} = 1 - \frac{6\sum {(Y - \sum_{n=0}^{6} \hat{y_n})}^2}{n(n^2 - 1)} \label{eq:5}
\end{equation}
We report SCC averaged across folds for the LOSO cross-validation scheme.
The intermediate OSATS scores are averaged into a signal-level OSATS score after processing the whole kinematic sample. The final six OSATS scores are then summed to give a video-level GRS. Note that the predicted GRS, \(\hat{Y} = \sum_{n=0}^{6} \hat{y_n} \), is only used to assess model performance, but is not directly learned from our model.

The mean average error (MAE) is also reported.
To our knowledge, Benmansour \textit{et al.} \cite{Benmansour2023} is the only recent work reporting OSATS-specific performance under the LOSO validation scheme. OSATS performance is reported at the same epoch used to report GRS performance under the same training parameters. OSATS SCC was averaged across 10 epochs. 

\noindent \textbf{Data Augmentation}: Two augmentation techniques were added to the kinematic signals to improve generalisation: 1) Gaussian noise based on the standard deviation of the signal, and 2) flipping, i.e. reversing the signal. Augmentations were done at a rate of \(50\%\). Label smoothing and dropout was performed at 30\%.

\noindent \textbf{Implementation}:
Our model is trained with the Adam optimizer for 5000 epochs with a learning rate of \(10^{-6}\). Kinematic data was pre-processed by normalising across time and feature dimensions~\cite{mikel}. The kinematics from the slave and master device were used (\(D = 76\)).
The sequence length of 75 was chosen. It corresponds to 2.5s in time (force data acquired at 30hz) and is consistent with the minimal time required for our clinician to rate a gesture. A lambda of 0.01 for L2 regularization and a batch size of 25 was used.
In line with existing literature, design decisions and hyperparameter adjustments were experimentally conducted using the averaged cross-validation test fold~\cite{Evangelos}. ReCAP was implemented in Pytroch and trained on an Nvidia A100 GPU.

\noindent \textbf{Validation of Model Behaviour}:
To validate our model's ability to generate interim OSATS scores, we asked a consultant surgeon in endoscopic interventions to agree or disagree with the model's intermediate predictions for the OSATS of Overall Performance. Every 75 frames was assigned a generated pseudo-label. The label is shown on the screen during viewing.
Similar to Wang \textit{et al.}'s framework \cite{Wang2020}, the surgeon was aware of the ground truth i.e. the trial level OSATS label.
The predicted OSATS scores were divided into three categories: poor (1-2), average (3), and good (4-5).
We randomly generate some segment predictions in two videos to mitigate potential bias without informing the surgeon. We then present these predictions and capture agreement or disagreement at the segment level when playing the video sequentially. 
\begin{table}[!htbp]
\centering
\begin{minipage}{0.49\textwidth}
\centering
    \caption{Perofrmance for OSATS scores, where the $\rho Osats$ is the average across the 6 scores under LOSO scheme. *: results from training across the 3 tasks. Across Tasks (AT)}
        \label{tab:OSATS average}
        \begin{tabular}{|c|c|c|c|c|}
            \hline
            & KT & NP & SU & AT \\
            \hline
            Apen\cite{Zia2018} & 0.66 & 0.45 & 0.59 & 0.57\\
            \hdashline
            FCN \cite{IsmailFawaz2019} & 0.65 & \textbf{0.57} & 0.60 & \textbf{0.61}\\
            \hdashline
            \textbf{ReCAP} & \textbf{0.70} & 0.46 & \textbf{0.62} & 0.59/0.58*\\
            \hline
        \end{tabular}
\end{minipage}
\begin{minipage}{0.49\textwidth}
\centering
    \caption{Ablation of ReCAP components for GRS under LOSO scheme.}
        \label{tab:ablation}
        \begin{tabular}{|c|c|c|c|}
            \hline
            & KT & NP & SU \\
             \hline
            ReCAP no augmentation & 0.86 & 0.85 & 0.83 \\
            \hdashline
            ReCAP no pseudo-label & 0.85 & 0.54 & 0.28 \\
            \hdashline
            \textbf{ReCAP} & \textbf{0.88} & \textbf{0.85} & \textbf{0.83}\\
            \hline
        \end{tabular}
\end{minipage}
\end{table}
\begin{table}[!htbp]
\label{tab:scoresspecific}
        \centering
        
        \caption{OSATS performance,$\rho Osats$ is reported for RT: \textit{Respect for tissue}, TM: \textit{Time and Motion}, OP: \textit{Overall Performance}, SNH:  \textit{Suture and Needle Handling}, FO: \textit{Flow of Operation}, QFP: \textit{Quality of Final Product} for the best/average fold performance. Across tasks(AT) is trained on the three tasks.\cite{Benmansour2023} only report best results }
        \label{tab:OSATS Specific Results}
        \begin{tabular}{|c|c|c|c|c|c|c|c|}
        \hline
        \multicolumn{1}{|c|}{} &\multicolumn{3}{c|}{CNN+Bilstm \cite{Benmansour2023}} & \multicolumn{4}{c|}{\textbf{ReCAP}} \\
        \hdashline
         \multicolumn{1}{|c|}{} & 
        \multicolumn{1}{c}{KT} & 
         \multicolumn{1}{c}{NP} &  \multicolumn{1}{c|}{SU}& 
        \multicolumn{1}{c}{KT} & 
         \multicolumn{1}{c}{NP} &  \multicolumn{1}{c|}{SU} &
         \multicolumn{1}{c|}{AT}\\
        \hline
        RT & \multicolumn{1}{c}{0.83} & \multicolumn{1}{c}{0.49} & \multicolumn{1}{c|}{0.46} &  \multicolumn{1}{c}{\textbf{0.92}/0.78} & \multicolumn{1}{c}{\textbf{0.75}/0.43} & \multicolumn{1}{c|}{\textbf{0.78}/0.52} & \multicolumn{1}{c|}{0.56} \\
        \hline
        TM & \multicolumn{1}{c}{0.87} & \multicolumn{1}{c}{0.85} & \multicolumn{1}{c|}{0.68} &  \multicolumn{1}{c}{\textbf{0.95}/0.8} & \multicolumn{1}{c}{\textbf{0.91}/0.72} & \multicolumn{1}{c|}{\textbf{0.84}/0.60} & \multicolumn{1}{c|}{0.62} \\
        \hline
        OP & \multicolumn{1}{c}{0.89} & \multicolumn{1}{c}{\textbf{0.58}} & \multicolumn{1}{c|}{0.71} &  \multicolumn{1}{c}{\textbf{0.9}/0.79} & \multicolumn{1}{c}{0.42/0.23} & \multicolumn{1}{c|}{\textbf{0.69}/0.5} & \multicolumn{1}{c|}{0.65} \\
        \hline
        SNH & \multicolumn{1}{c}{0.82} & \multicolumn{1}{c}{0.79} & \multicolumn{1}{c|}{0.75} &  \multicolumn{1}{c}{\textbf{0.84}/0.61} & \multicolumn{1}{c}{\textbf{0.91}/0.69} & \multicolumn{1}{c|}{\textbf{0.88}/0.78} & \multicolumn{1}{c|}{0.65} \\
        \hline
        FO & \multicolumn{1}{c}{0.76} & \multicolumn{1}{c}{0.58} & \multicolumn{1}{c|}{0.62} &  \multicolumn{1}{c}{\textbf{0.78}/0.63} & \multicolumn{1}{c}{\textbf{0.66}/0.45} & \multicolumn{1}{c|}{\textbf{0.89}/0.66} & \multicolumn{1}{c|}{0.64} \\
        \hline
        QFP &  \multicolumn{1}{c}{0.75} & \multicolumn{1}{c}{0.31} & \multicolumn{1}{c|}{0.67} &  \multicolumn{1}{c}{\textbf{0.85}/0.59} & \multicolumn{1}{c}{\textbf{0.56}/0.22} & \multicolumn{1}{c|}{\textbf{0.91}/0.64} & \multicolumn{1}{c|}{0.62} \\
        \hline
        AVG &  \multicolumn{1}{c}{0.82} & \multicolumn{1}{c}{0.60} & \multicolumn{1}{c|}{0.65} &  \multicolumn{1}{c}{\textbf{0.87}/0.70} & \multicolumn{1}{c}{\textbf{0.70}/0.46} & \multicolumn{1}{c|}{\textbf{0.83}/0.62} & \multicolumn{1}{c|}{0.62} \\
        \hline
        \end{tabular}
\end{table}

\section{Results}

We report the performance of our model against previous work that uses kinematic data or video data and report GRS performance (Table \ref{tab:results}). Although we don't regress the GRS's most recent work only report on it. To allow for easier comparison we use the GRS as a performance proxy. The model outperforms all methods using kinematic data and achieves competitive performance against models using video (Table \ref{tab:results}). 
When looking at the performance of our model in predicting OSATS under the LOSO validation scheme, we underperform only in NP (Table \ref{tab:OSATS average}).
The CNN+Bilstm \cite{Benmansour2023} only reports the best-performing fold, whereas we also report the average across the 5 folds.
\begin{table}[htbp]
\centering
    \caption{Performance comparison of GRS score on JIGSAWS trained on independent and across tasks. \textbf{K}: Kinematic, \textbf{V}: Video. *: Results from training on the three domains. GRS is used as a performance proxy and not regressed directly in this paper.}
\label{tab:results}
\begin{tabular}{|c|c|c|c|c|c|}
 \hline
 \multicolumn{2}{|c|}{} & \multicolumn{4}{c|}{} \\
 \multicolumn{2}{|c|}{} & \multicolumn{4}{c|}{Task}\\
\multicolumn{1}{|c}{Input} & \multicolumn{1}{c|}{Method} &\multicolumn{4}{c|}{} \\
\cline{3-6}
 \multicolumn{2}{|c|}{} & \multicolumn{1}{c|}{KT} & \multicolumn{1}{c|}{NP}& \multicolumn{1}{c|}{SU}& \multicolumn{1}{c|}{AT}\\
 \hline
 \multicolumn{2}{|c|}{} &  \multicolumn{4}{c|}{\textbf{Spearman's Correlation Coef (SCC)}} \\
 \hdashline
 \multicolumn{1}{|c}{\textbf{V}} & \multicolumn{1}{c|}{C3D-MTL-VF \cite{IMTL}}  & \multicolumn{1}{c}{\underline{\textit{0.89}}} & \multicolumn{1}{|c}{0.75} & \multicolumn{1}{|c}{0.77} & \multicolumn{1}{|c|}{0.80} \\
 \hdashline
 \multicolumn{1}{|c}{\textbf{V}} & \multicolumn{1}{c|}{Contra-Sformer \cite{Evangelos}}  & \multicolumn{1}{c}{\underline{\textit{0.89}}} & \multicolumn{1}{|c}{0.71} & \multicolumn{1}{|c}{\textbf{0.86}} & \multicolumn{1}{|c|}{0.82} \\
 \hdashline
 \multicolumn{1}{|c}{\textbf{V}} & \multicolumn{1}{c|}{ViSA \cite{Li}}  & \multicolumn{1}{c}{\textbf{0.92}} & \multicolumn{1}{|c}{\textbf{0.93}} & \multicolumn{1}{|c}{\underline{\textit{0.84}}} & \multicolumn{1}{|c|}{\textbf{0.90}} \\
\hline
\multicolumn{1}{|c}{\textbf{K}} & \multicolumn{1}{c|}{SMT-DCT-DFT\cite{Zia2018}}  & \multicolumn{1}{c}{0.70} & \multicolumn{1}{|c}{0.38} & \multicolumn{1}{|c}{0.64} & \multicolumn{1}{|c|}{0.59} \\
\multicolumn{1}{|c}{\textbf{K}} & \multicolumn{1}{c|}{DCT-DFT-ApEn\cite{Zia2018}}  & \multicolumn{1}{c}{0.63} & \multicolumn{1}{|c}{0.46} & \multicolumn{1}{|c}{0.75} & \multicolumn{1}{|c|}{0.63} \\
 \hdashline
\multicolumn{1}{|c}{\textbf{K}} & \multicolumn{1}{c|}{\textbf{ReCAP}} & \multicolumn{1}{c}{0.88} & \multicolumn{1}{|c}{\underline{\textit{0.85}}} & \multicolumn{1}{|c}{0.83} & \multicolumn{1}{|c|}{\textit{\underline{\textit{0.85}}}/0.79*} \\
\hline
\multicolumn{2}{|c|}{} &  \multicolumn{4}{c|}{\textbf{Mean Average Error (MAE)}} \\
 \hdashline
 \multicolumn{1}{|c}{\textbf{V}} & \multicolumn{1}{c|}{Contra-Sformer \cite{Evangelos}}  & \multicolumn{1}{c}{\textbf{1.75}} & \multicolumn{1}{|c}{3.15}  & \multicolumn{1}{|c}{2.74}  & \multicolumn{1}{|c|}{2.55} \\
\hdashline
\multicolumn{1}{|c}{\textbf{V}} & \multicolumn{1}{c|}{ViSa \cite{Li}}  & \multicolumn{1}{c}{2.16} & \multicolumn{1}{|c}{\textbf{1.66}} & \multicolumn{1}{|c}{\textbf{2.58}} & \multicolumn{1}{|c|}{\textbf{2.13}} \\
\hdashline
\multicolumn{1}{|c}{\textbf{K}} & \multicolumn{1}{c|}{\textbf{ReCAP}}  & \multicolumn{1}{c}{2.04} & \multicolumn{1}{|c}{3.12} & \multicolumn{1}{|c}{2.89} & \multicolumn{1}{|c|}{2.68/2.71*} \\
\hline
\end{tabular}
\end{table}
As can be seen in Table \ref{tab:ablation} the introduced guassian noise and flipping had very little effect on the performance of the model. However the the flipping does allow for the model to be time invariant. We see that the pseudo-label drastically improves performance, especially for the two tasks, NP and SU, with the most class imbalance~\cite{Lefor2020}.  

To validate the weakly supervised outputs, 9 videos were reviewed by a consultant surgeon, where each 75 frames (the segment length) had an assigned OSATS pseudo-label. The selected videos regrouped three levels of expertise (novice, intermediate, expert) across the three tasks. Two of those videos were shown with the randomly generated predictions. We found that the clinician agreed 69\% of the time when shown random predictions while agreeing 77\% of the time when shown our model's predictions. A one-tailed binomial test between the two distributions indicates a statistically significant difference between the agreements (p=0.006). 
\begin{figure*}
    \centering
    \includegraphics[width=1\columnwidth]{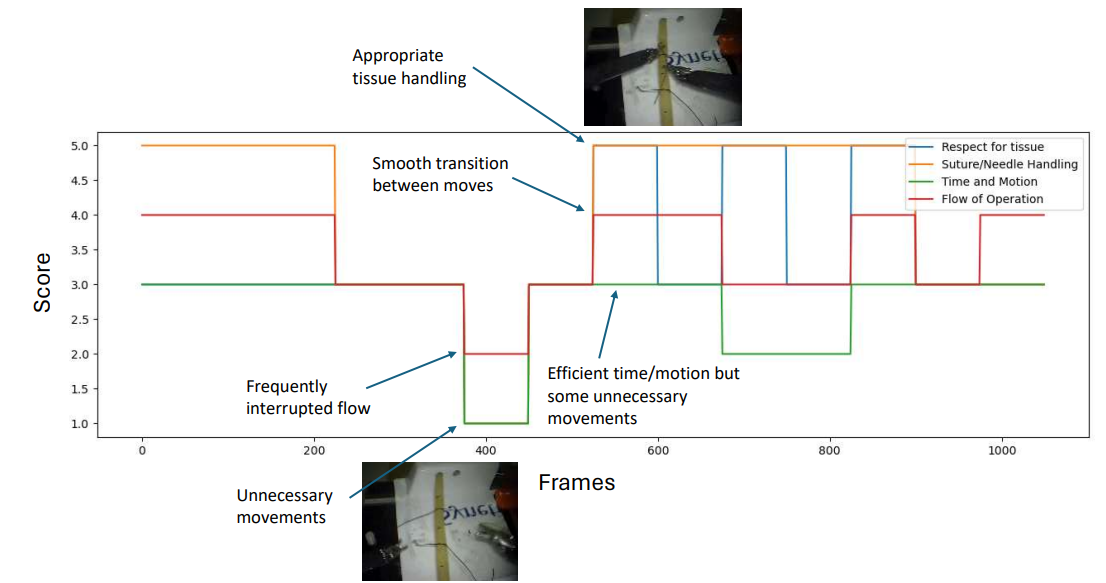}
    \caption{Variations of OSATS from model for a Knot Tying task performed by a self-designated expert. Qualitative descriptions are taken from \cite{Martin1997}. In this example, the senior clinician disagreed with the model's intermediate scores in 3 instances.}
    \label{fig:pipeline}
\end{figure*}
\section{Discussion}
To our knowledge we are the first to report task-agnostic metrics on JIGSAWS. While video-based models dominate recent works; kinematic data, though underexplored, remains promising. Our model outperforms prior kinematic-only methods and rivals video-only approaches, despite a simple architecture. The high feature-to-parameter ratio (238,644:1,440) contributes to overfitting, common in deep learning.

Our model shows strong OSATS prediction overall, though underperforms on Needle Passing and Quality of Final Product. This likely stems from kinematic data's inability to capture visual nuances. As noted by Kasa \textit{et al.}~\cite{Kasa2022}, quality of final product is very video subjective. Similar kinematic profiles may yield different outcomes—e.g., a depth misjudgment in Needle Passing affects performance but not kinematics.

As Lefor \textit{et al.}~\cite{Lefor2020} noted, JIGSAWS is imbalanced, with some OSATS scores inversely correlated with skill level. Also, using Spearman’s $\rho$ on only 3 samples for certain folds is very biased. Thus, model generalisation to real-world practice remains limited.

Our model benefits from a formulation that segments input with temporal context, enabling sparse, flexible learning. Ablations show that adding pseudo-labels improves performance, likely by regularising via intermediate predictions. This mirrors human raters, who assess cumulatively. However, the current objective doesn’t capture catastrophic errors well; allowing segment-wise weighting could address this.

Pseudo-label loss aids both performance and interpretability. We visualise this in Fig.~\ref{fig:pipeline}. It supports clinician feedback and online use due to the model’s recurrent nature. Our work is limited by validation of fine-graned OSATS scores, which are difficult to distinguish by experts, as seen in the 69\% agreement with random noise rather than the expected 33\%. Rater variability complicates ground truth extraction. While more data and raters would help, it’s often impractical. Weak supervision offers a scalable alternative. Our model's performance vs. noise suggests promise in this direction.

\section{Conclusion and Future Work}
We present a novel formulation for skill assessment, extensible to recurrent models and other domains. Our competitive results on JIGSAWS support its potential for granular feedback. Future work will improve validation, incorporate more OR time-series data (audio, bodytracking, ...)~\cite{julien}, and target longer, complex procedures. Since annotating multi-hour tasks is laborious, weakly-supervised methods extracting gesture/step/phase-level labels may enable scalable, automated assessment.

\bibliographystyle{splncs04}
\bibliography{Paper-0006}

\end{document}